\documentclass{article}

\PassOptionsToPackage{numbers,sort&compress}{natbib}
\usepackage[dblblindworkshop, final]{neurips_2026}
\makeatletter
\renewcommand{\section}{\@startsection{section}{1}{\z@}%
  {-1.5ex \@plus -0.3ex \@minus -0.2ex}
  {0.8ex \@plus 0.2ex}
  {\large\bf\raggedright}}
\renewcommand{\subsection}{\@startsection{subsection}{2}{\z@}%
  {-1.2ex \@plus -0.3ex \@minus -0.2ex}%
  {0.5ex \@plus 0.2ex}%
  {\normalsize\bf\raggedright}}
\makeatother

\usepackage[utf8]{inputenc} 
\usepackage[T1]{fontenc}    
\usepackage{hyperref}       
\usepackage{url}            
\usepackage{booktabs}       
\usepackage{amsfonts}       
\usepackage{amsmath}        
\usepackage{nicefrac}       
\usepackage{microtype}      
\usepackage{xcolor}         
\usepackage{multirow}       
\usepackage{graphicx}       
\usepackage{rotating}       
\usepackage{soul}           
\usepackage{xcolor-material}

\hypersetup{pdftitle={Are Seismic Foundation Models More Than Glorified Feature Extractors?}, pdfauthor={Laura Laurenti, David Liu, Brian Kulis, Men-Andrin Meier, Christopher Johnson}}

\newif\ifshownotes\shownotesfalse
\ifshownotes
  \DeclareRobustCommand{\todo}[1]{{\color{MaterialRed900}\textbf{[TODO: #1]}}}

\else
  \DeclareRobustCommand{\todo}[1]{}
\fi

\title{Are Seismic Foundation Models More Than Glorified Feature Extractors?}

\workshoptitle{AI for Stochastic Dynamics}

\author{%
  \hspace*{-\tabcolsep}Laura Laurenti$^{1,2*}$, David Liu$^{3*}$, Brian Kulis$^{4}$, Men-Andrin Meier$^{1,2}$, Christopher Johnson$^{5}$\hspace*{-\tabcolsep} \\
  \smash{$^{1}$}Institute of Geophysics, ETH Zurich \quad \smash{$^{2}$}Swiss Seismological Service, ETH Zurich \\
  \smash{$^{3}$}Dept.\ of Computer Science, \smash{$^{4}$}Dept.\ of Electrical and Computer Engineering, Boston University \\
  \smash{$^{5}$}Los Alamos National Laboratory \\
  \smash{$^{*}$}Equal contribution \quad Corresponding author: \texttt{laura.laurenti@sed.ethz.ch} \\
}

\begin{document}

\maketitle

\begin{abstract}
Earthquake sequences are stochastic on two separated
timescales: the seismic waves of any individual earthquake propagate over
seconds to minutes, while the earthquakes themselves arrive as a sparse,
self-exciting occurrence process, with inter-event intervals from
seconds to months. Respecting both scales in one training objective is the open
challenge. Foundation models (FMs) for seismic waveforms instead import their learning
targets from audio and vision, where every target is computed from a single
fixed-length window. Such a window holds one realization of the wave propagation
(an earthquake trace), never the occurrence process that decides when
earthquakes happen. We develop nine model designs sharing one Mamba
encoder--decoder, trained with a fixed corpus of ${\sim}3$M windows, and we vary
the self-supervised objective, the quantization bottleneck and the mixer: the
two comparisons we build on are architecture-matched, varying the objective
alone. Each checkpoint is evaluated frozen on three downstream
tasks. Varying the target redistributes downstream skill rather than raising it, and no checkpoint wins everywhere. 
Adding masking pressure to our strongest classifier re-specializes it rather than making it a generalist: it gains $6.9$ points 
of few-shot P-wave picking precision (detecting when the first seismic wave arrives) and loses $9.6$ points of few-shot classification 
accuracy. By the standard imported from language and vision, these models are
feature extractors rather than FMs. Controls on capacity,
architecture, and domain proximity do not explain why skill moves rather than rises.
Against task-specific baselines the advantage is regime-dependent: our checkpoints 
beat them few-shot on classification and on ground motion (how strongly the ground will shake), 
and trail them on full data. We argue the binding constraint is the target, and that settling the
question requires a learning target defined across events, the timescale that
makes seismicity stochastic in the first place.
\end{abstract}



\section{Introduction}
\label{sec:intro}
Seismicity is a canonical stochastic dynamical system at two
distinct levels. \emph{Within} a recorded earthquake trace, ground motion is a
continuous,
non-stationary process governed by the wave equation \citep{boore2003stochastic}, 
with randomness from source and path heterogeneity, site response, and ambient noise 
\citep{bonnefoyclaudet2006noise}. \emph{Across} the catalog (the record of past events with their times, locations and magnitudes), 
earthquakes arrive as a sparse, self-exciting point process: inter-event intervals span seconds
during an aftershock sequence to months during interseismic quiescence
\citep{utsu1995omori}, and
magnitudes follow the Gutenberg--Richter law \citep{gutenberg1944frequency}, so
the events that matter most (large magnitude events) are by construction the rarest. The modeling tradition for the second level, like ETAS
\citep{ogata1988statistical} and its neural descendants
\citep{mei2017neuralhawkes, zuo2020transformerhawkes, dascher2023deepetas},
is explicitly a stochastic-dynamics tradition.
Characterizing a sequence means saying when the next event will come and how large it will be: quantities that live in the catalog, not in any one trace.

Machine learning has entered at the first level, the waveforms themselves.
These are continuous ground-motion recordings from networks of seismometers irregularly placed at the surface, 
cut into fixed-length windows of three orthogonal components (vertical and two horizontal), typically sampled at $100$ or $200$\,Hz.
Recording never stops, but only the sparse windows around catalogued events carry labels, and the 
curated benchmarks used for pretraining are smaller still: on the order of a 
million traces \citep{mousavi2019stead, suarez2024crew}.
Recent foundation models (FMs) for seismic waveforms \citep{liu2024seislm, si2024seisclip, li2024seist,
utrans2026, wang2026seismollm, cai2026seismicxm, cheng2025gsfm, laurenti2026}
import self-supervised recipes from language, vision, and audio domains
\citep{devlin2019bert, he2022mae, baevski2020wav2vec}. These recipes differ in
architecture and objective, but agree on something more basic and rarely
stated: the learning target is always a functional of a single
fixed-length window. Masked prediction \citep{liu2024seislm}, compressed reconstruction \citep{laurenti2026}, contrastive alignment \citep{si2024seisclip}, and diffusion denoising \citep{cheng2025gsfm} all ask the model to recover or discriminate one realisation of the observation process, on a fixed clock; none references the occurrence dynamics, which bounds the abilities of the FM. When the physics is local to a short window, 
a window-level target is enough, and phase picking is the clear case, except when 
signal of interest is the pattern of occurrence (Section~\ref{sec:interp}).

We ask what that choice costs, under conditions tight enough to attribute the
answer: corpus, heads and splits fixed throughout, and, for the two comparisons
we build on, architecture fixed too with the objective varied alone. The
finding is negative, and we argue it is diagnostic. Varying the window-level
target \emph{redistributes} downstream skill rather than raising it. Controls 
on capacity, architecture, and domain proximity do not explain the
ceiling. That points at the target without saying \emph{which} property of it is
at fault, so we distinguish throughout what we demonstrate from what we
motivate.

\section{Setup}
\label{sec:setup} 

We pretrain nine models on one shared backbone: a SEANet-style \citep{tagliasacchi2020seanet} convolutional encoder--decoder whose
every stage is paired with a block of Mamba selective state-space layers
\citep{gu2023mamba}, operating on five-minute, three-component $100$\,Hz
traces. Mamba is chosen over attention for linear-time scaling at these lengths; SEANet is an audio codec, 
so the backbone is itself one of the imports we examine. It is held fixed, to isolate the target 
rather than the architecture. Details on the backbone are in Appendix~\ref{app:backbone}. These are small models by FM standards,
$2.8$--$6.0$M parameters, hence the $57$M scaling control (Section~\ref{sec:results},
with the recipe in Appendix~\ref{app:controls}).
Pretraining draws on CREW \citep{suarez2024crew}, ${\approx}1.6$M regional
earthquake waveforms, paired batch-for-batch with an ambient-noise corpus we
assembled from $462$ FDSN networks (${\approx}1.5$M training windows), so the
model reconstructs one pure seismic noise window per earthquake window. Appendix~\ref{app:corpora} gives full detail about the corpora.

The nine checkpoints span three axes, not one. \emph{Objective}: reconstruction
(\textit{Rec}) composed with a magnitude head (\textit{Mag}), an adversarial
discriminator (\textit{Disc}), and span masking (\textit{Mask}).
\emph{Bottleneck}: a continuous latent, or residual vector quantisation
(\textit{RVQ}) updated by backpropagation or EMA. \emph{Mixer}: SSM-only, or SSM
plus a gated-MLP branch (\textit{+MLP}). Two sub-comparisons are architecture-matched, and they
are the ones the argument rests on: the four \textit{Rec}-family rows share one
architecture and differ only in loss composition, and \textit{+MaskRaw} and
\textit{+MaskCode} match \textit{RVQ-EMA+MLP} in state-dict shape, adding only a
training-time objective that is discarded at inference. See Appendix~\ref{app:objectives} 
for full objectives detail. A seismic FM could in principle serve the whole monitoring pipeline
\citep{woollam2022seisbench} and the study of how seismicity 
evolves \citep{laurenti2024, rouetleduc2017mlquakes}. We attempt three of these tasks, 
listed below.\\
Each checkpoint is \emph{frozen} and evaluated with lightweight heads on three
tasks (see Appendix~\ref{app:protocol} for more details about the downstream tasks): foreshock/aftershock (earthquakes before or after the largest event of a sequence) classification at station NRCA around the 2016
$M6.5$ Norcia mainshock \citep{laurenti2024}; regression of five peak
ground-motion measures for early warning over a 39-station Central-Italy network
\citep{bloemheuvel2022gcn, Jozinovic2020}; and detection of P/S arrival times (picking the arrival of fast primary and slower secondary waves) \citep{zhu2019phasenet} on STEAD 
\citep{mousavi2019stead}. Each is scored
full-data and few-shot (${\approx}100$ labels) against the published
task-specific model trained directly on it \citep{laurenti2024,
bloemheuvel2022gcn, zhu2019phasenet}. The encoder sees none of the three downstream corpora during pretraining, so every downstream task is a cross-corpus generalisation test: what we report is 
specialisation \emph{under distribution shift}, the regime FMs are meant to serve.

\section{Results}
\label{sec:results}

\begin{table}[t]
\centering
\footnotesize
\setlength{\tabcolsep}{4pt}
\caption{Frozen-encoder downstream evaluation, full-data and few-shot
(${\approx}100$ labelled examples). \textit{FEQ/AEQ}: foreshock/aftershock
accuracy; \textit{PGA}: peak-ground-acceleration MSE, reported for both
ground-motion heads we train (a GCN over a station-distance graph and an
independent-station linear head); \textit{P-/S-prec}: picking precision at
$0.5$\,s tolerance. Baselines are the external task-specific models; the
ground-motion baseline is itself a GCN, so it is matched by our GCN column and
has no linear-head counterpart in the literature (dashes).
\textbf{Bold} marks the best value in a column, \underline{underline} our best
checkpoint; no ordering survives across tasks.}
\label{tab:main}
\begin{tabular}{@{}lcccccccccc@{}}
\toprule
\multirow{2}{*}{\textbf{Objective}}
 & \multicolumn{2}{c}{\textbf{FEQ/AEQ} (\%) $\uparrow$}
 & \multicolumn{2}{c}{\textbf{PGA} GCN $\downarrow$}
 & \multicolumn{2}{c}{\textbf{PGA} linear $\downarrow$}
 & \multicolumn{2}{c}{\textbf{P-prec} (\%) $\uparrow$}
 & \multicolumn{2}{c}{\textbf{S-prec} (\%) $\uparrow$} \\
\cmidrule(lr){2-3}\cmidrule(lr){4-5}\cmidrule(lr){6-7}\cmidrule(lr){8-9}\cmidrule(lr){10-11}
 & full & few & full & few & full & few & full & few & full & few \\
\midrule
Task-specific baseline & \textbf{99.2} & 82.4 & 0.20 & 0.331 & --- & --- & \textbf{96.5} & \textbf{86.8} & 93.8 & 87.7 \\
\midrule
Rec                & 87.2 & 50.6 & 0.344 & 0.328 & 0.338 & 0.326 & 93.0 & 18.8 & 93.5 & 66.1 \\
Rec+Mag            & 77.8 & 63.8 & 0.198 & 0.299 & 0.213 & 0.288 & 89.5 & 45.3 & 92.5 & 74.4 \\
Rec+Disc+Mag       & 93.9 & 85.4 & 0.230 & 0.306 & 0.197 & 0.285 & 93.9 & 68.8 & 93.9 & 82.9 \\
Rec+Disc+Mag+Mask  & 87.2 & 82.9 & \textbf{\underline{0.161}} & \textbf{\underline{0.268}} & \textbf{\underline{0.162}} & \textbf{\underline{0.263}} & 95.8 & \underline{83.5} & \textbf{\underline{94.2}} & \textbf{\underline{88.1}} \\
RVQ                & 93.9 & 75.6 & 0.248 & 0.308 & 0.199 & 0.280 & 94.9 & 70.1 & 93.9 & 81.3 \\
RVQ-EMA            & 95.6 & 70.3 & 0.243 & 0.323 & 0.193 & 0.293 & 95.1 & 72.9 & 94.0 & 83.7 \\
RVQ-EMA+MLP        & \underline{96.7} & \textbf{\underline{85.5}} & 0.306 & 0.330 & 0.197 & 0.305 & 95.5 & 74.1 & 93.9 & 84.1 \\
RVQ-EMA+MaskRaw    & 93.6 & 75.9 & 0.221 & 0.315 & 0.174 & 0.281 & \underline{96.1} & 81.0 & 94.0 & 88.0 \\
RVQ-EMA+MaskCode   & 96.3 & 81.2 & 0.269 & 0.326 & 0.187 & 0.296 & 95.8 & 79.3 & 94.1 & 82.5 \\
\bottomrule
\end{tabular}
\end{table}

No target wins everywhere: objective composition dominates
Table~\ref{tab:main}, and dominates \emph{differently} per task.
\textit{RVQ-EMA+MLP} is our best classifier in both regimes ($96.7$, $85.5$) yet
is beaten on ground motion and S-picking by \textit{Rec+Disc+Mag+Mask} (PGA
$0.161$ vs.\ $0.306$ full-data), which in turn trails it by $9.5$ points
full-data. The effect is large: adding adversarial, magnitude and masking
terms to reconstruction raises few-shot P-precision from $18.8$ to $83.5$,
and larger than any architectural change we tested. Baselines are likewise
task-dependent: our best checkpoint clears the task-specific CNN \citep{laurenti2024} few-shot
($85.5$ vs.\ $82.4$) but trails it full-data ($96.7$ vs.\ $99.2$), so
pretraining helps where labels are scarcest and not otherwise. Ground motion
replicates this on the head that admits a comparison: against the GCN baseline \citep{bloemheuvel2022gcn} every checkpoint wins on all five targets few-shot, whereas full-data only
\textit{Rec+Disc+Mag+Mask} does, the rest on two of five at best, and four of
them on none. On picking the two heads are close, splitting by wave: 
the baseline \citep{zhu2019phasenet} slightly ahead on P ($0.4$ points full-data, $3.3$ few-shot), ours slightly 
ahead on S ($\le0.4$, seven of nine full-data and our best few-shot). There is no best checkpoint here, only a best checkpoint for a stated task and label budget.

As the controlled test, we add to our strongest classifier
(\textit{RVQ-EMA+MLP}) the masking ingredient that helps within the
reconstruction family, on the hypothesis that more self-supervised signal is
generically useful. It does not produce a generalist: \textit{+MaskRaw} gains
$6.9$ points of few-shot P-precision ($74.1\rightarrow81.0$) and our best
full-data P-precision ($96.1$), while losing $9.6$ points of few-shot
classification ($85.5\rightarrow75.9$). \textit{+MaskCode}, whose masking
touches only a side branch discarded at inference, trades less both ways, as
expected if a target shapes the representation to the extent it applies pressure
upstream of the encoder. Adding self-supervised signal therefore moves
specialisation rather than dissolving it: specialisation is a property of
\emph{what the target asks the model to represent}, not of how much signal it
receives.

Three controls,
scored on classification, rule out the readings that do not implicate the
target. All three are our own runs from a larger set of pretraining variations
not reported here.
\emph{Capacity}: width-scaling the encoder ${\sim}11\times$
(encoder+decoder $4.96\rightarrow56.8$M) recovers only part of a corpus-induced deficit, so the ceiling is not a small-model
artefact. \emph{Architecture}: bidirectional Mamba mixers, a better inductive bias in
principle here, change nothing measurable at roughly double the mixer cost. \emph{Domain
proximity}: adding $1.2$M domain-matched local traces leaves both metrics tied,
so the limit is not a wrong-distribution corpus. Throughout, the full-data split
is insensitive to almost everything we vary; few-shot is both the discriminating
and the fragile metric.

\section{Analysis and open problems}
\label{sec:interp}

Capacity, architecture, and domain proximity are ruled out above, leaving the
target. Two properties of it are in play: we demonstrate the first and motivate
the second, and mark the difference. We then discuss four problems, all about the target rather than the architecture.

\textbf{Demonstrated: a point target on a stochastic process is
misspecified.} In a further experiment (Appendix~\ref{app:forecast}) we asked the
model to forecast $12$\,s of ground motion under an $L_1/L_2$ loss. It output a
flat line, $5$--$10\times$ below the true spectrum where the energy is. This is
the optimum, not a failure to reach it: squared error is minimised by the
conditional mean $\mathbb{E}[Y \mid X]$, and $12$\,s ahead the phase of the
horizon waveform is nearly independent of the context, so that mean is
${\approx}0$. A perfectly trained model is therefore flat, and the loss cannot
separate it from one that has learned the physics. The repair is to target something 
predictable: multi-scale RMS-envelope agreement, which is what early warning needs anyway. 
A point target silently asks for a conditional mean, 
so it teaches only where that mean is predictable \citep{song2021sde}.
Downstream the ordering then inverts: the collapsed arm yields features
${\sim}4.5$ points \emph{better} than the envelope arm whose forecasts look
right. What a target buys is what it forces the encoder to retain, not the
sample it emits.

\textbf{Motivated but untested: the single-window target of Section~\ref{sec:intro}}. A target computed from a single window can only describe what is inside
it. The occurrence dynamics are not: the inter-event interval, the triggering
relation to a mainshock months earlier, and the clustering statistics
operational forecasting relies on \citep{ogata1988statistical} are all discarded
before the loss is evaluated. This is the gap between modelling one sample path
and modelling the law that generates it, and where our work and the neural-SDE
literature \citep{kidger2021neuralsde} diverge. We cannot yet show this is the
binding constraint: every objective we varied and every task we scored is
window-level, so the redistribution is equally consistent with ordinary task
conflict at fixed capacity. Separating the two needs an occurrence-level target (discussed 
below in Problem~1). Figure~\ref{fig:mismatch} illustrates the three structural 
differences behind this paragraph and the problems below.

\textbf{(P1) Two-timescale learning targets.} An objective could couple a
waveform-level encoder to a marked temporal point process over the catalogue, so
that one target sees both the within-trace dynamics and the between-event
occurrence dynamics. The neural point-process literature
\citep{mei2017neuralhawkes, zuo2020transformerhawkes, dascher2023deepetas}
already models occurrence. What is missing is a formulation in which the
waveform representation and the event history condition each other. Such a target would let waveform content sharpen occurrence forecasts, including the prospective version of the foreshock question we can currently only ask after the fact.

\textbf{(P2) Tail-aware generative targets.} Diffusion and score-based models 
\citep{ho2020ddpm, song2021sde} are already viable on seismic-adjacent signals 
\citep{cheng2025gsfm}, but have been applied to signal processing rather than 
event-centric inference. Given the conditional-mean collapse above, the 
question is not whether the target should be distributional: it is how to weight the 
objective so that rare events contribute gradient signal 
in proportion to how much they matter, not their frequency.

\textbf{(P3) Evaluation with a stated cost structure.} No rule exists for
weighing a $9.6$-point classification loss against a $6.9$-point picking gain,
so which checkpoint to deploy is underspecified within this study, let alone
across methods. A benchmark for learned stochastic simulators needs proper
scoring rules \citep{gneiting2007scoring}, calibration targets, and an explicit
cost structure for cross-task trade-offs. With those, a result like ours reads
as a Pareto frontier rather than a null finding. Seismic FMs are 
new enough that this benchmark can still be designed rather than inherited.

\textbf{(P4) Targets that use the known physics.} Wave propagation has a known
governing operator, so a target need not rediscover it: defining the objective
on the residual to a forward model, or in a neural-operator basis, would spend
capacity only on the stochastic part, that is, what the physics leaves
undetermined. Both components already exist (neural operators \citep{rahman2022uno, rahman2022gano} 
and physics-residual training \citep{smith2021eikonet}) as ways to solve the physics, 
not as pretraining targets.

\section{Conclusion}
We asked whether seismic FMs are more than feature extractors. To answer, 
we developed nine checkpoints sharing one backbone and one corpus, and we evaluated them with a frozen encoder on 
three tasks under distribution shift. These are competent pretrained feature extractors for seismology rather 
than FMs in the sense imported from language and vision. Varying the objective redistributes 
downstream skill rather than raising it, and no checkpoint generalises without giving something up; 
pretraining helps where labels are scarce (few-shot training). Controls on capacity, architecture and domain proximity do not explain 
why skill moves rather than rises, so what remains is the target.
We demonstrate one of its failures, the collapse of a point target to a 
near-zero conditional mean, and can so far only motivate the other, that a single-window target cannot 
reference the occurrence process. What stochastic theory contributes is therefore not an architecture 
but a criterion for when a learning target is well posed.

\newpage

\bibliographystyle{plainnat}
\bibliography{mybibfile}

\newpage

\appendix
\renewcommand{\thefigure}{A\arabic{figure}}

\section{Full experimental detail}
\label{app:setup}

\subsection{Backbone}
\label{app:backbone} A SEANet-style encoder--decoder: an \texttt{SConv1d +
ResnetBlock} front-end followed by three strided downsampling stages (strides
$3,4,5$), each paired with a block of Mamba selective state-space layers
\citep{gu2023mamba}, and a symmetric decoder. Mamba is chosen over attention for
linear-time scaling at these sequence lengths ($3\times30{,}000$ samples per
window, reduced $60\times$ along time to ${\sim}500$ latent frames). Where present, the RVQ bottleneck is $8\times1024\times128$; the
\textit{+MLP} setting enables a gated-MLP branch with $d_{\text{intermediate}} =
2 d_{\text{model}}$. Downstream heads always consume the continuous
pre-quantisation representation, never the discrete codes.

\subsection{Objectives}
\label{app:objectives} \textit{Rec} combines $L_1/L_2$ losses on the raw
waveform (weight $5$ each) and on the one-sided amplitude spectrum
($\nicefrac{1}{3}$, $\nicefrac{1}{100}$). \textit{Disc} is a hinge
adversarial loss (weight $\nicefrac{1}{3}$) with an $L_1$ feature-matching term
over intermediate discriminator features, the latter carrying a large effective
weight ($100/3$) that matters below. \textit{Mag} is an auxiliary
magnitude-regression head (MSE on the predicted mean plus a small penalty on the
log-variance).
\textit{Mask} applies span masking to the raw input (ratio $0.4$, span $40$
samples), with the waveform term restricted to masked positions;
\textit{+MaskCode} instead masks a separate copy of the encoder output (ratio
$0.5$, span $10$ latent frames) consumed only by an auxiliary cross-entropy head
predicting the first three RVQ levels ($0.83$M parameters, discarded at
inference).

\subsection{Corpora}
\label{app:corpora} CREW \citep{suarez2024crew}: ${\approx}1.6$M five-minute,
three-component regional traces ($1$--$20^\circ$). Ambient noise: $1{,}499{,}918$
training windows (${\approx}125{,}000$ hours) from $462$ FDSN networks, plus
$169{,}832$ validation and $162{,}846$ test windows. Waveforms are
peak-normalised jointly across the three components, preserving inter-channel
amplitude ratios.

\subsection{Downstream protocol}
\label{app:protocol} Foreshock/aftershock \citep{laurenti2024}: two months pre- and
post-mainshock windows ($25$\,s) at NRCA station, $70/10/20$ split
full-data and $2\%$ few-shot. Ground motion \citep{Jozinovic2020,bloemheuvel2022gcn}: $10$\,s post-origin waveforms,
$915$ events $\times$ $39$ INGV stations, predicting PGA, PGV, and
PSA$_{0.3/1.0/3.0}$ through both a 2-layer GCN head over a station-distance
graph and an independent-station linear head; $65\%$ full-data, $8\%$ few-shot.
Phase detection \citep{zhu2019phasenet}: per-sample 3-way (noise/P/S) labelling on the SeisBench STEAD
partition, $143{,}523$ training traces, a prediction counted correct within
$50$ samples ($0.5$\,s) of the labelled arrival; $0.1\%$ few-shot. All numbers
in Table~\ref{tab:main} use a frozen encoder.

\subsection{Scale, architecture and corpus controls}
\label{app:controls} 
The three controls quoted in Section~\ref{sec:results}, drawn from a larger set of pretraining variations, are reported here only in brief.\\
Width scaling raises the base channel width $16\rightarrow56$ (encoder+decoder $4.96\rightarrow56.8$M;
with the RVQ codebooks, $6.0\rightarrow57.9$M, the convention used in
Section~\ref{sec:results}),
holding depth, downsampling ratios, window length, batch size and learning rate
fixed; this architecture scales by width, not depth, since per-stage Mamba depth
$56$ makes the backward pass effectively hang. Its few-shot figures are quoted
against a same-corpus $6$M control rather than against Table~\ref{tab:main},
because the scaling runs use a pooled CREW\,+\,INSTANCE \citep{michelini2021instance} \,+\,STEAD corpus and a
shorter $6000$-sample window. Bidirectional Mamba replaces every mixer with a
forward-plus-reversed scan fused by a linear projection, at ${\approx}2\times$
mixer cost, on the same CREW-plus-noise corpus as Table~\ref{tab:main}. 
Corpus rebalancing compares two mixtures at identical architecture, 
both drawn from STEAD, INSTANCE, SCEDC and ETHZ with CREW capped at 
${\approx}\nicefrac{1}{10}$: a proportional mixture, 
and a domain-first mixture adding $1.2$M further locally recorded traces.

\subsection{The forecasting experiment}
\label{app:forecast} 
Separately from the nine checkpoints in Table~\ref{tab:main}, this experiment 
uses a domain-first mixture of ${\approx}3.35$M traces (STEAD, 
INSTANCE, SCEDC and ETHZ, with CREW capped at ${\approx}\nicefrac{1}{10}$ of the total) 
cropped to a common $6000$-sample window because the shortest constituent corpus is $60$,s. 
Each window splits into a $4800$-sample context and a genuinely held-out $1200$-sample horizon. The 
horizon is never an encoder input, which distinguishes this from the masking
variants. A latent forecast head (20 learnable future queries rolled forward by
a 2-layer causal Mamba) predicts a future latent feeding both the shared decoder
and a next-token head over the first three RVQ levels. Arm~1 applies $L_1/L_2$
plus a spectra term to the predicted horizon waveform; arm~2 replaces the two
time-domain terms with multi-scale RMS-envelope agreement over
$W=\{25,50,100,200\}$ samples ($0.25$--$2$\,s), keeping the already
phase-invariant spectra term. Both are run against a matched control (identical
model, data, seed and windowing, forecast terms removed) over 3 seeds. Arm~1
reaches $86.1\pm1.0$ few-shot classification against arm~2's $81.6\pm0.9$ (the
${\sim}4.5$-point gap quoted in Section~\ref{sec:interp}) while producing the
collapsed forecast described there. As with the scaling control, these figures
are comparable to each other and to their own control, not to
Table~\ref{tab:main}: the matched control reaches $78.9$ few-shot against
Table~\ref{tab:main}'s $85.5$, and that entire gap is the data recipe rather
than the objective.

\begin{sidewaysfigure}
\centering
\includegraphics[width=0.95\textheight]{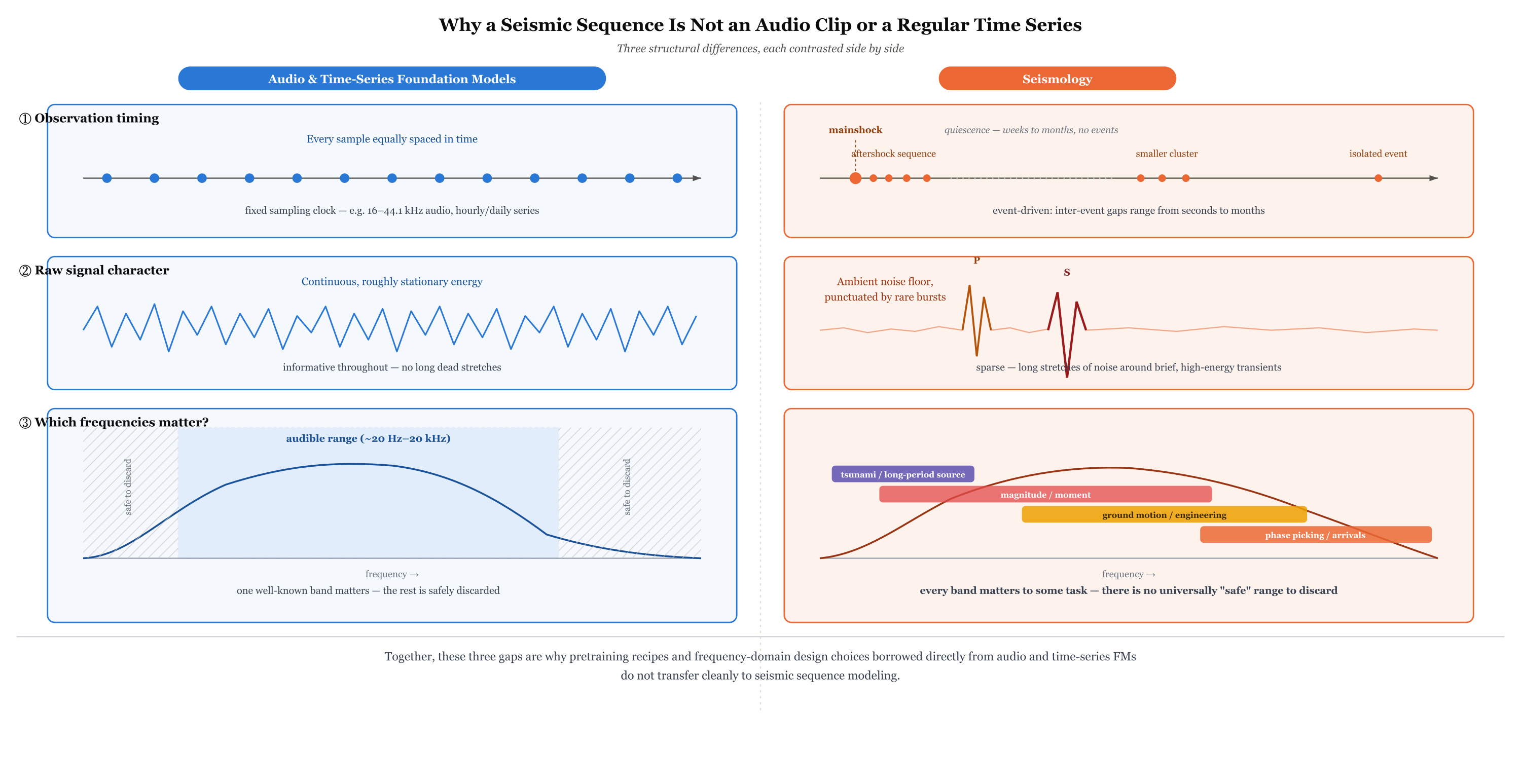}
\caption{The three structural differences between the signals audio and
time-series FMs are built for and the sequences seismology cares
about. \textbf{(1)} Audio and forecasting corpora sit on a fixed
sampling clock, whereas earthquake occurrence is event-driven with inter-event
gaps from seconds to months, the reason a window-level target cannot reference
the occurrence process. \textbf{(2)} Audio carries roughly stationary energy
throughout; seismic traces are a noise floor punctuated by rare, high-energy
transients: the asymmetry that makes an $L_1/L_2$ objective mis-weighted for a
heavy-tailed law. \textbf{(3)} Audio codecs can design around a known audible
band and discard the rest; seismology has no universally safe band, since
picking, magnitude, and engineering applications each depend on a different part
of the spectrum, so any compression target pre-selects which downstream tasks
survive it.}
\label{fig:mismatch}
\end{sidewaysfigure}

\end{document}